\documentclass[10pt,twocolumn,letterpaper]{article}

\usepackage{cvpr}              % To produce the CAMERA-READY version
\usepackage{graphicx}
\usepackage{amsmath}
\usepackage{amssymb}
\usepackage{booktabs}

\usepackage{times}  % DO NOT CHANGE THIS
\usepackage{helvet}  % DO NOT CHANGE THIS
\usepackage{courier}  % DO NOT CHANGE THIS
\usepackage[hyphens]{url}  % DO NOT CHANGE THIS
\usepackage{caption} % DO NOT CHANGE THIS AND DO NOT ADD ANY OPTIONS TO IT
\usepackage{pifont}
\DeclareCaptionStyle{ruled}{labelfont=normalfont,labelsep=colon,strut=off} % DO NOT CHANGE THIS
\usepackage{multirow}
\usepackage{setspace}
\usepackage{bbding}
\usepackage{epsfig}
\usepackage{float}
\usepackage{color}
\usepackage{ulem}

\newcommand{\hupar}[1]{\vspace{12pt} \noindent \textbf{#1}\quad}
\newcommand{\tabincell}[2]{\begin{tabular}{@{}#1@{}}#2\end{tabular}}

\usepackage[pagebackref,breaklinks,colorlinks]{hyperref}

\usepackage[capitalize]{cleveref}
\crefname{section}{Sec.}{Secs.}
\Crefname{section}{Section}{Sections}
\Crefname{table}{Table}{Tables}
\crefname{table}{Tab.}{Tabs.}

\def\cvprPaperID{5331} % *** Enter the CVPR Paper ID here
\def\confName{CVPR}
\def\confYear{2022}

\begin{document}

\title{Optical Flow Estimation for Spiking Camera}
\author{Liwen Hu\textsuperscript{\rm 1,2}\thanks{These authors contributed equally to this work.}, 
Rui Zhao\textsuperscript{\rm 1*}, 
Ziluo Ding\textsuperscript{\rm 1}, 
Lei Ma\textsuperscript{\rm 1,2}\thanks{Corresponding author.},
Boxin Shi\textsuperscript{\rm 1,2,3},
Ruiqin Xiong\textsuperscript{\rm 1}, 
Tiejun Huang\textsuperscript{\rm 1,2,3}\\
\textsuperscript{\rm 1}NERCVT, School of Computer Science, Peking University\\
\textsuperscript{\rm 2}Beijing Academy of Artificial Intelligence\\
\textsuperscript{\rm 3}Institute for Artificial Intelligence, Peking University\\
%{\tt\small yj.zheng@pku.edu.cn, zhenglingxiao@pku.edu.cn, yuzf12@pku.edu.cn, shiboxin@pku.edu.cn, \\ yhtian@pku.edu.cn, tjhuang@pku.edu.cn}
}
% 我们需要一个更明确的名字以及缩写，后面再讨论 for --> with ? 
% High - Speed Optical Flow Estimation with xxx? 

\maketitle

\begin{abstract}
%As a bio-inspired sensor with high temporal resolution, the spiking camera has an enormous potential in real applications, especially for motion estimation in high-speed scenes. However, frame-based and event-based optical flow methods are not well matched to the spike stream from the spiking camera due to the different data modalities. In this paper, we presents, SCFlow, a novel deep learning pipeline to perform optical flow estimation for the spiking camera, the basic task in motion estimation.
%the spiking camera that mimics the retina fovea can report per-pixel luminance intensity accumulation by firing spikes asynchronously. 
As a bio-inspired sensor with high temporal resolution, the spiking camera has an enormous potential in real applications, especially for motion estimation in high-speed scenes. However, frame-based and event-based methods are not well suited to spike streams from the spiking camera due to the different data modalities. To this end, we present, SCFlow, a tailored deep learning pipeline to estimate optical flow in high-speed scenes from spike streams. Importantly, a novel input representation is introduced which can adaptively remove the motion blur in spike streams according to the prior motion.
% 话题的转折有点突然 scflow有着动态的编码模块，能适应性地根据先验运动将被给的脉冲流转化成合适的输入表征。
Further, for training SCFlow, we synthesize two sets of optical flow data for the spiking camera, SPIkingly Flying Things and Photo-realistic High-speed Motion, denoted as SPIFT and PHM respectively, corresponding to random high-speed and well-designed scenes. Experimental results show that the SCFlow can predict optical flow from spike streams in different high-speed scenes. Moreover, SCFlow shows promising generalization on \textbf{real spike streams}. Codes and datasets refer to \url{https://github.com/Acnext/Optical-Flow-For-Spiking-Camera}.

%\textit{All codes and constructed datasets will be released after publication}.
% \lei{Please re-re-emphasize our contribution more clearly. I have made some adjustments, which is far from enough. Remember: after reading abstract, reviewer has decided his final score 30 -percent. Delete this in the final version. }
\end{abstract}
%提纲： 要不要改成实时应用 real app 改成 real-time app
%引言：1 光流的重要性 传统相机的不足2 事件相机的优势与不足 脉冲相机的好处 与不足3 我们的贡献 
%remarkable sole
%相关工作：1 脉冲相机 2 光流方法(图像域 事件域 脉冲域) 这个要修改一下了 改为1 image-based 光流方法 2 event-based 3 spike-based 然后单独开一章简单讨论脉冲相机原理和做光流的意义与挑战
%Discussion on the spiking camera 1 脉冲相机原理 2脉冲相机上的光流估计 问题设置 挑战
%方法：1模拟器2网络模型
%实验：1 实验配置(数据集 训练细节 等等) 2 性能对比(其实最好跟event camera作比较。因为动作变化不大，spike转image再比较可能还是没啥差距，而event数据则会大大丢失信息肯定不如spike光流。如果真的和image比较不如确定好帧率关系，比如spike用40000fps的数据，image用120fps的数据。此外可以看看最早的event光流文章和谁比的)
%结论
\section{Introduction}
%As a method to evaluate the motion of objects, optical flow estimation has always been a hot topic in computer vision and it is widely applied on moving object segmentation [], human pose estimation [], and action recognition []. However, for high-speed scenes, the accuracy of optical flow estimation is greatly limited by the blurred images (as fig1(b)) sampled by traditional cameras with low frame rate. To better estimate the optical flow in high-speed scenes, some work [] begin to directly extract optical flow from events stream sampled by event cameras \cite{dvs1,dvs2,dvs3,dvs4,dvs5}. Although event cameras with high temporal resolution can avoid the motion blur, the event stream (as fig) that only encoding the change of luminance intensity is also insufficient for optical flow estimation in some scenes, especially in low texture scenes.

Optical flow estimation has always been a popular topic in computer vision and played important roles in a wide range of applications, such as object segmentation \cite{2018OpticalFlowForSegmentation}, video enhancement \cite{wang2020deep}, and action recognition \cite{tu2019action}. However, the breakthrough of this field in high-speed scenes is impeded by blurry images from traditional cameras with low frame rate.
%Although traditional cameras with high frame rates can avoid motion blur to a certain extent, they are not suitable for being widely used in real applications due to their high cost.
%尽管有着高帧率的传统相机能在一定程度上避免运动模糊，它们不适合广泛地被使用到实际应用中由于昂贵的造价。
%这句话是不是有点踩dvs，要不要换种说法看看别人文章有没有提这点 然后cite下。
\begin{figure}[ht]
\includegraphics[width=\linewidth]{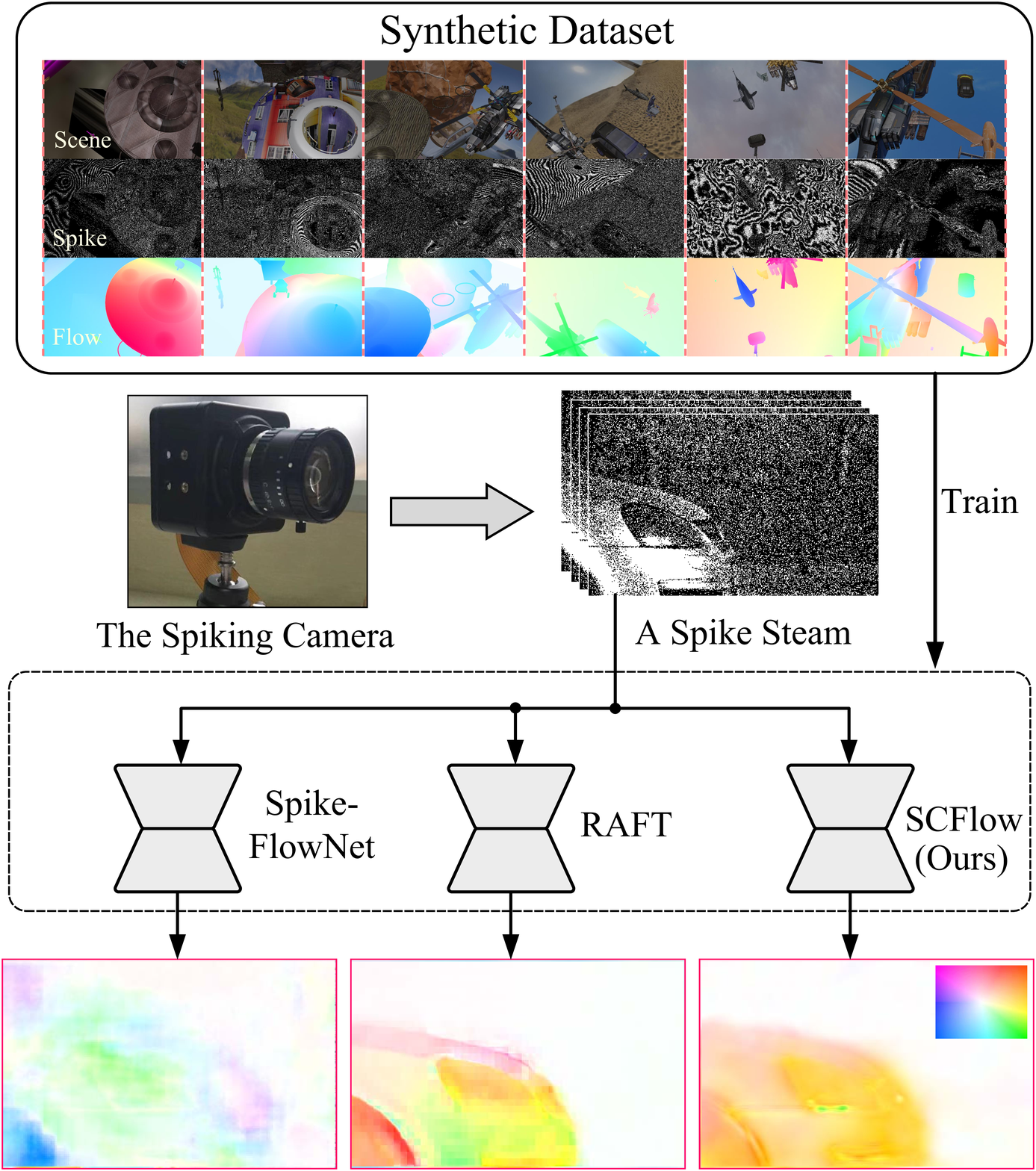}
\centering
\caption{The optical flow estimation for \textbf{a real spike stream} which records a car traveling at a speed of 100 km/h. We compare our SCFlow to event-based method (Spike-FlowNet \cite{lee2020spike}) and frame-based method (RAFT \cite{teed2020raft}). \textbf{All methods use spike streams as input and are trained on proposed dataset (SPIFT)}. Results show that our proposed method has better performance which clearly distinguishes regions with different motion and accurately predict optical flow in each region. On the top right corner of results is a visualization of the color coding of the optical flow.%\lei{Figure 1 should preferably correspond closely to introduction. " Tell your story!"} \hu{Yes, the fig1 can reflect our contribution but can not the whole story, like motivation. I was wondering if we need  to epress the story since papers I saw only reflect own contribution}
}\label{fig0}
\end{figure}
%Different from event cameras, the spiking camera \cite{spikecode, spikecamera} can report per-pixel luminance intensity accumulation by firing spikes asynchronously. For each sample, if luminance intensity accumulation in a pixel arrives predefined threshold, a spike (represented by '1') can be sent and the accumulation is reset, otherwise there is no spike (represented by '0') in the pixel. Hence, instead of grayscale image, the output of all pixels forms a binary matrix (spike frame) and all spike frames form spike stream. Benefiting from the sampling method, the spiking camera not only has high temporal resolution (40000Hz), but also records the texture details of objects meaning sampled scenes, even high-speed scenes, can be reconstructed directly from spike data with high quality \cite{spikecamera, rec1, rec2, rec3}. Hence, the coded information in spike stream is sufficient for optical flow estimation in high-speed scenes. However, the classical image-based and event-based methods cannot be directly applied to binary spike stream.
%\\\indent(event camera \cite{dvs1,dvs2,dvs3,dvs4,dvs5} and the spiking camera \cite{spikecode, spikecamera})
The emergence of neuromorphic cameras \cite{dvs1,dvs2,dvs3,dvs4,dvs5, spikecode, spikecamera} provides a new perspective for optical flow estimation in high-speed scenes.
Some works \cite{zhu2018multivehicle,zhu2019unsupervised,lee2020spike} raise the interest in event cameras \cite{dvs1,dvs2,dvs3,dvs4,dvs5} and show optical flow in high-speed scene can be directly estimated from an event stream.
%展示了用事件流提取光流有好处
However, the event stream that only encodes the change of luminance intensity might be insufficient for optical flow estimation in all regions of a scene, especially for regions with weak textures. Also as a neuromorphic camera, the spiking camera \cite{spikecode, spikecamera} not only has high temporal resolution (40000Hz) but can  report per-pixel luminance intensity by firing spikes asynchronously.
Specifically, each pixel in the spiking camera can accumulate incoming light independently and persistently. At each timestamp, if luminance intensity accumulation at a pixel exceeds the predefined threshold, a spike is fired and the accumulation is reset for that pixel, otherwise there is no spike at that position. Hence, instead of a grayscale image, the output of all pixels forms a binary matrix representing the presence of spikes, also known as a  spike frame, and continuous spike frames form a spike stream. Further, the sampled high-speed scene can be reconstructed from the spike stream\cite{spikecamera, rec1, rec2, rec3, rec4}. Hence, the spiking camera that can record details of objects has an enormous potential for optical flow estimation in high-speed scenes. %\lei{Please try to rewrite this paragraph in the order shown in Figure 1, perhaps with some 'a','b','c' subfigure}
\\\indent
At present, there is no research about spike-based optical flow estimation, one of the challenges is that the spike stream has a unique data modality so that frame-based and event-based methods are not % suitable for
directly applicable to it. For estimating optical flow from spike stream, an intuitive solution is to reconstruct image sequences from spike stream firstly, and then use frame-based methods to estimate optical flow. 
However, when the spike stream over a period of time is converted into a two-dimensional image, there is a time offset between the reconstructed image and the real scene which would bring additional errors to optical flow estimation.
Besides, simple reconstruction methods \cite{spikecamera, rec1, rec2} are difficult to filter out the motion blur in the spike stream while high-quality reconstruction methods \cite{rec3, rec4} would cost a lot of extra processing efforts. %\lei{This is not enough, I believed the lost of time accuracy is much more vital. That is is to say, 1. delay. 2.the reconstruction requires a accumulation of multiple spikes, which leads to inaccuracy of time. } %手工方法 能这么说么?
%简单(手工？)方法难以达到高质量的重构由于容易遭受噪声影响 而 使用深度学习的高质量重构将花费大量的额外时间. 更重要地
Therefore, it is necessary to design a tailored method to estimate optical flow directly from spike streams. Another challenge is there are no optical flow datasets for the spiking camera to properly evaluate the performance of spike-based optical flow methods. In fact, it is difficult to build real optical flow datasets for the spiking camera since calibrating ground truth optical flow is chanllenging in high-speed scenes \cite{de2020evaluating, pandey2011ford}. Hence, synthetic spiking optical flow datasets seem to be the more feasible  way to solve this challenge.

In this paper, we propose SCFlow, a neural network tailored to estimate optical flow directly from spike streams. Different from previous work using deep learning \cite{rec3, rec4} where the spike stream in temporal windows with fixed direction is used as features, we propose a novel input representation for spike streams, Flow-guided Adaptive Window (FAW). By adaptively selecting temporal windows for each pixel based on the prior motion, FAW can avoid the motion blur \cite{spikecamera} in the spike stream caused by static temporal windows. Besides, for training our network and evaluating the performance, we synthesize two spike-based optical flow datasets, SPIkingly Flying Things and Photo-realistic High-speed Motion, denoted as SPIFT and PHM respectively.

%SPIFT and PHM include 110 random scenes and 10 well-designed scenes respectively. Random scenes in SPIFT comprise random objects that translate and rotate in random background, like FlyingChairs \cite{dosovitskiy2015flownet}. Well-designed scenes in PHM are more lifelike and has a lot in common with the real world. \lei{move such detailed descriptions to "related work", unless we want to emphasize the 'datasets' contribution. }%

%Empirically, w
We show that SCFlow can estimate optical flow accurately in high-speed scenes and achieve the state-of-the-art performance in comparison with existing frame-based and event-based methods on our datasets. {Importantly}, SCFlow shows promising generalization on real spike streams as shown in Fig.~\ref{fig0}. %First, we introduce a novel image-based representation of a given spike stream where (write key point of representation)Second, (introduce network)Finally, the first optical flow dataset for the spiking camera is generated by our proposed the spiking camera simulator which combining simulation function and rendering engine tightly. {Dataset name} includes 50 scenes as training set and 10 scenes as testing set. The training set includes random objects that translate and rotate in random background, like Flying Chair[]. The testing set is carefully designed and has a lot in common with the real world.By estimating in dataset, we show that SCFlow can predict optical flow well in high-speed scenes and achieve the state-of-the-art performance comparing to image-based and event-based methods.
%原有表征: here the frequencies of spike in different time windows are extracted
%Recently, deep learning has shown great potential
%in vision applications [31, 10, 24, 31, 10, 24].
%In this paper, we propose a novel neural network architecture, SCFlow, to estimate the optical flow from spike stream. First, the spikes in temporal windows are generally used as features for all pixels \cite{rec3}. However, the temporal windows should be carefully chosen at each pixel due to the potential light change and object movement. Thus, we propose a novel input representation of spike stream which allows temporal windows to be adaptively selected based on the prior motion.  %(重点看看)

% 第一个问题要不要说单图像+事件提取光流，如果说那么他的缺点是信息不充分吗，感觉不是，单pixel+事件也能复原很好的场景。
%第二个问题 scenes与 scenarios的用法
%第三个问题 frame-based 和 image-based区别 frame-based可能是针对于相机说的我都改成了image-based 因为 vidar也是frame-based
% 一些工作which尝试利用模糊图像+脉冲数据重建出图像序列再做光流提取 取得了更好的效果，但是额外的转换本身就牺牲了大量的时间与空间。 不同与事件相机，模仿视网膜中央凹的脉冲相机[]can report per-pixel luminance intensity accumulation by firing 被1bit表示的 spikes。 得益于他的采样原理, the spiking camera[] not only has high temporal resolution, but also records the texture details of objects， and 我们能够使用特定算法[重构文章]直接从脉冲数据复原出高质量的采样场景。 因此, 脉冲数据携带的信息是充足的to 提取高速场景下物体的光流。然而，图像域和事件域的光流估计无法直接在脉冲数据上生效。as a result, there has been a significant research drive to develop new 光流估计算法 suitable for the spiking camera。
%第三段写贡献，从网络开始写 可以展开讲

In general, we attempt to exploit the potential of the spiking camera in high-speed motion estimation and our main contributions are summarized as follows:

\begin{itemize}
\item[1)] %We are the \emph{first} work to explore optical flow estimation for the spiking camera, and propose a novel neural network architecture where a novel input representation, FAW, allowing adaptive temporal window selection is used to cope with the motion blur of spike stream in temporal windows with fixed direction.
We propose the first work to explore optical flow estimation in high-speed scenes with the spiking camera, and propose a tailored neural network architecture with a novel input representation, FAW, for handling the motion blur in a spike stream in temporal windows with fixed direction. %\lei{can we be so sure about two "novel" in the first contribution?}

\item[2)]
%表征： where The predicted optical flow at the previous time is used as guidance
We synthesize the first spike-based optical flow datasets (SPIFT and PHM) to benchmark optical flow estimation for the spiking camera, which includes well-designed scenes with various motion%. In addition to the task, we hope our datasets could 
, and to inspire future research on spike-based vision tasks.
%这个which后面讲这个表征的作用感觉更好
\item[3)]
%We train the SCFlow on proposed datasets, experimental results demonstrate that our proposed SCFlow can estimate flow field from spike stream efficiently. Moreover, SCFlow shows promising generalization on \emph{real} spike stream.
We demonstrate that SCFlow can estimate flow field from the spike stream on proposed datasets efficiently. Importantly, SCFlow can be generalized well on real spike streams captured in real high-speed scenarios.
%\lei{generalized well?}%\lei{It's better to mention that scflow can handle ultra-high-speed scenes, otherwise it can't see the advantage of doing of with "image reconstruction and then OF". It is to correspond to the challenges mentioned above.} 
\end{itemize}

% \begin{itemize}
% \item[1)]
% We generate the \textit{first} spike-based optical flow datasets, SPIFT and PHM by proposed the spiking camera simulator (SPCS). 
% \end{itemize}
% \begin{itemize}
% \item[2)]
% %表征： where The predicted optical flow at the previous time is used as guidance
% We present SCFlow, the \textit{first} deep learning architecture for optical flow estimation of the spiking camera. Importantly, we introduce a novel input representation, FAW, allowing adaptive temporal window selection, which is fed into SCFlow as the sole input.
% %这个which后面讲这个表征的作用感觉更好
% \end{itemize}
% \begin{itemize}
% \item[3)]
% We train SCFlow on SPIFT dataset and estimate optical flow for the spiking camera both on synthetic dataset and real dataset. Experimental results demonstrates that our proposed SCFlow can estimate flow field from spike stream efficiently.
% \end{itemize}

% \lei{Please re-re-emphasize our contribution more clearly as well. Remember: after reading contributions, reviewer has decided his final score 50-percent. Delete this in the final version. }

% 总的来说，我们的目的是充分发掘the巨大潜力of光流估计from脉冲相机and我们的贡献可以总结如下:
%1 首次提出了脉冲相机的模拟器which紧密结合了模拟功能和渲染引擎允。通过模拟脉冲相机的采样原理，模拟器可以在我们搭建的场景中生成相应的脉冲数据同时其他用于视觉任务的数据e.g.,也能被生成。
%2 。进一步，我们搭建了各种复杂的运动场景包括(), and, 通过使用这个模拟器，我们提出了首个脉冲光流数据集。 (有点少看看怎么加)
%3 一个xx网络被提出 （介绍一下特点）通过在数据集上训练，我们发现使用脉冲数据训练的xx网络相比framebase上的xx和eventbase上的xx有更好的性能and达到了sota。

%相关工作
\section{Related Work}
\subsection{Frame-based and Event-based Optical Flow}
Optical flow estimation for frame-based cameras has been a classical vision task since it was first introduced by Horn and Schunck \cite{horn1981determining}. Early methods describe the essence of flow field via an illumination consistency assumption and combine it with a smoothness constraint to avoid the ill-posed condition. Many effective modules were introduced to subsequent algorithms such as estimating the flow fields coarse-to-fine via a pyramid structure and warping \cite{brox2004high} and median filtering \cite{sun2010secrets}. However, these variational methods suffer a huge time cost. In the variational age, the datasets to evaluate an optical flow algorithm are mainly Middlebury \cite{baker2011database}, Sintel \cite{butler2012naturalistic} and KITTI \cite{geiger2012we, menze2015object}. The flow ground truth of Middlebury dataset is obtained via UV illumination or artifical synthesis, which has only dozens of samples. The Sintel dataset is derived from an open source 3D animated short film. The KITTI dataset gets flow ground truth with LIDAR, which causes the flow field  to be sparse. However, these datasets are not adequate in quantitive terms for training a deep neural network.

Synthesizing data from computer graphics model has been shown effectiveness in computer vision \cite{richter2016playing}.
Dosovitskiy et al. \cite{dosovitskiy2015flownet} firstly propose a large dataset FlyingChairs to train an end-to-end neural network FlowNet via supervised learning. FlowNet 2.0 \cite{ilg2017flownet} improves the performance by stacking the network, which leads to the oversize of the model.  Knowledge from classical methods such as pyramid, warping \cite{brox2004high} and cost volume \cite{scharstein2002taxonomy} were introduced to optical flow estimation network to make it compact \cite{sun2018pwc}. However, ground truth of optical flow is hard to obtain. Deep optical flow networks trained by unsupervised scheme were proposed to handle this problem \cite{jason2016back}, similar with variational methods, it employs photometric loss and smoothness loss. To improve the reliability of supervision signal, bi-directional flow estimation \cite{meister2018unflow} was proposed to detect occlusion area and stop its back-propagation. Recently, self-supervision methods \cite{liu2019selflow, liu2020learning} were proposed to improve the performance of unsupervised networks.
Optical flow estimation for event cameras has attracted more and more interest due to its high temporal resolution. The MVSEC \cite{zhu2018multivehicle} dataset gets flow ground truth through LIDAR and records natural scenes using event cameras and gray cameras simultaneously. EV-FlowNet \cite{zhu2018ev} can be regarded as the first deep learning method for event-based optical flow, which is trained by photometric loss and smoothness loss with the help of gray images. Zhu et al. \cite{zhu2019unsupervised} train the network with a loss function designed to eliminate the motion blur in event streams. SpikeFlowNet \cite{lee2020spike} propose a hybrid network with spiking neural network (SNN) encoder to better exploit the temporal information in event streams.
% , imporving the performance in long time range conditions. 
STEFlow \cite{ding2021spatio} further improves the performance using recurrent neural networks as its encoder.

\subsection{The Spiking Camera and Its Applications}%简单介绍原理和应用 不用描述model
% the spiking camera is a bio-inspired sensor with high temporal resolution. Different from event cameras, the spiking camera \cite{spikecode, spikecamera} can report per-pixel luminance intensity accumulation by firing spikes asynchronously. Benefit from its distinct sampling mechanism, the spiking camera can record all the texture details of objects \textit{theoretically}. However, the DAVIS \cite{dvs3} may need other image sensor for supplementary texture information. %q:这句的表达可能有点问题
% Given its huge potential in applications, especially for high-speed scenes, low-level vision tasks based on the spiking camera have developed rapidly. \cite{spikecamera} first provided high-speed imaging for the spiking camera by counting the time interval (TFI) and the number of spikes (TFP). \cite{rec0} improved the smoothness of reconstructed images through motion aligned filtering.
% \cite{rec1, rec2} and \cite{rec3} respectively use SNN and convolutional neural network to reconstruct high-speed images from spike stream, which greatly improved the reconstruction quality.
The spiking camera is a bio-inspired sensor with high temporal resolution. Different from event cameras, it \cite{spikecode, spikecamera} can report per-pixel luminance intensity by firing spikes asynchronously. Benefiting from its distinct sampling mechanism, texture details of objects in high-speed scenes can be recorded theoretically.
Given its great potential in applications, especially for high-speed scenes, low-level vision tasks based on the spiking camera have developed rapidly. \cite{spikecamera} first reconstruct high-speed scenes by counting the time interval (TFI) and the number of spikes (TFP). \cite{rec0} improved the smoothness of reconstructed images through motion aligned filtering.
\cite{rec1, rec2} and \cite{rec3, rec4} respectively use SNN and convolutional neural network to reconstruct high-speed images from a spike stream, which greatly improved the reconstruction quality.
\cite{sup0} first present super-resolution framework for the spiking camera and recover external scenes with both high temporal and high spatial resolution from spike streams.
% problem：现在脉冲相机发展这块介绍的种类太少了 全是重构
%脉冲相机是一个有着高时间分辨率受生物启发的传感器。不同于事件相机, the spiking camera [] can report per-pixel luminance intensity accumulation by firing spikes asynchronously. Benefit from the sampling model, the spiking camera can directly record the texture details of objects without extra image sensor (DAVIS  []) or photo-measurement circuit (ATIS  [], CeleX  []). Hence, the camera has a huge potential in real-time applications and, recently, low level applications for the spiking camera [] have developed rapidly. [] 通过统计在脉冲流中的脉冲的时间间隔(TFI)与脉冲的数量(TFP)首次完成了脉冲相机的高速场景成像。[,] and [] 分别使用snn和ann对脉冲数据进行图像复原，大大提高了重建图像的质量。[]使用优化的方法完成了对脉冲流的超辨率图像重建。[] 首次实现了使用脉冲流进行光流估计。最后一句话需要补充一下
%%%%%%%%%%%%%%%%%%%%%%%%%%%%%%%%%%%%%
\section{Preliminary}
%In the section, the principle of the spiking camera is firstly introduced, and then the problem of optical flow estimation for the spiking camera and its challenges is discussed.
\subsection{The Spiking Camera Model}
\begin{figure}[htbp]
\includegraphics[width=\linewidth]{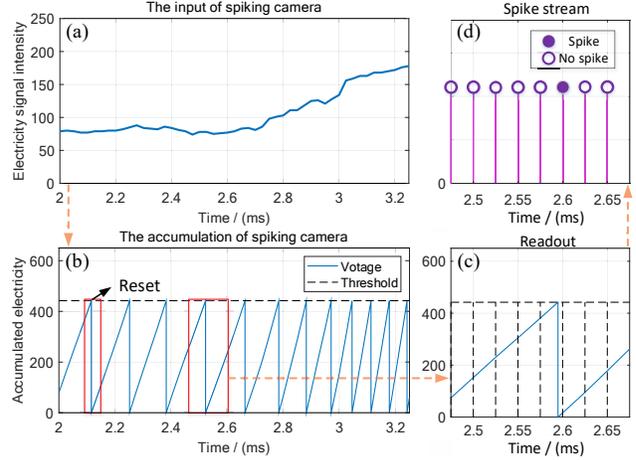}
\centering
\caption{Illustration of the spiking camera model. (a) The input electrical signal at a pixel. (b) The accumulation corresponding to the pixel. (c) The polling readout of spikes where the spiking camera triggers the generation of spike at a frequency of 40000 Hz. (d) Generated a spike stream at the pixel.}\label{123}
\end{figure}
The spiking camera mimicking the retina fovea consists of an array of $H \times W$ pixels and can report per-pixel luminance intensity  by firing spikes asynchronously. Specifically, each pixel on the spiking camera sensor accumulates incoming
light independently and persistently. At time  $t$, for pixel  $(i, j)$, if the accumulated brightness arrives a fixed threshold  $\phi$ (as (1)), then a spike is fired and the corresponding accumulator is reset as shown in Fig.~\ref{123}.
    %[]放一些中央凹的文章
   \begin{eqnarray}
    \mathbf{A}(i, j, t) = \int_{t_{i, j}^{\rm pre}}^{t} I(i, j, \tau) d\tau\geq \phi, 
    \end{eqnarray}
    where  $i, j\in \mathbb{Z}, i \leq H, j \leq W, k \leq N$,  $\mathbf{A}(i, j, t)$   is the accumulated brightness at time  $t$,  $I(i, j, \tau)$   refers to the brightness of pixel  $(i, j)$     at time  $\tau$, and   $t_{i, j}^{\rm pre}$   expresses the last time when a spike is fired at pixel  $(i, j)$  before time  $t$. If  $t$   is the first time to send a spike, then  $t_{i, j}^{\rm pre}$  is set as 0. In fact, due to the limitations of circuit technology, the spike reading times are quantified. Hence, asynchronous spikes are read out synchronously. Specifically, all pixels periodically check the spike flag at time $n\delta t, n \in \mathbb{Z}$, where $\delta t$  is a short interval of microseconds.
    Therefore, the output of all pixels forms a  $H \times W$   binary spiking frame. As time goes on, the camera would produce a sequence of spike frames, i.e., a  $H \times W \times N$  
binary spike stream and can be mathematically defined as, 
   \begin{eqnarray}
    \begin{aligned}
    &\mathbf{S}(i, j, n\delta t) =
    \\&
    \begin{cases}
    1 &\mbox{ if  $\exists t \in \left((n - 1)\delta t, n\delta t\right]$, s.t.  $\mathbf{A}(i, j, t) \geq \phi$   }, \\
    0 &\mbox{ if  $\forall t \in \left((n - 1)\delta t, n\delta t\right]$, $\mathbf{A}(i, j, t) < \phi$  },  \\
    \end{cases} 
    \end{aligned}
    \end{eqnarray}
    Accordingly, the average brightness of pixel  $(i, j)$    between times  $a$    and  $b$    can be evaluated by counting the number of spikes \cite{spikecamera}, \textit{i.e.},
    \begin{eqnarray}
     \sum\limits_{a \leq n\delta t \leq b} \!\!\!\! \mathbf{S}(i, j, n \delta t) \cdot \frac{\phi}{b - a}
    \end{eqnarray}
    
    %\lei{Please rewrite this formula. It is embarrassing and somehow unpleasant.. LOL.}
    
     %后面加一个相应地，一段时间内的平均光强能被使用xxx估计 (写出xxx这个表达式) 不再做别的讨论， 挑战的时候再说表征的问题 Accordingly, the average brightness of pixel  $(i, j)$    between time  $t$    and  $t_{i, j}^{\rm pre}$    can be calculated  \cite{spikecamera}, i.e., 
    %\begin{eqnarray}
    %\bar I(i, j)  \approx \dfrac{\phi}{t - t_{i, j}^{\rm pre}} = %\dfrac{\phi}{{n\Delta t}}, 
    %\end{eqnarray}
    %where,  $\Delta t$    is the sampling time interval and  $n \in \mathbb{N}$    denotes the number of intervals. 
%Benefiting from the sampling mechanism, the spiking camera has not only the similar advantage as event cameras, i.e. high temporal resolution (40000Hz), but also it can capture the texture detail of objects. Hence, ignoring the motion blur in traditional camera and the absence of texture in event camera, the spiking camera has natural advantages in optical flow estimation.
%

\subsection{Optical Flow for The Spiking Camera}
%\noindent\textbf{Problem Statement}
\hupar{Problem Statement.}We use  $S_n \in \{0, 1\}^{W \times H}$ to denote the $n$ spike frames and  $W_{t_i,t_j}$   to denote optical flow from time  $t_i$ to time $t_j$. %这个定义后期可以再讨论一下
 Given a recorded binary spike stream $\{S_n\}, n = 0,1,\cdots k$   from start to end of sampling, the goal of the spiking camera optical flow estimation is to predict the optical flow  $W_{t_i,t_j}$ based on the spike stream.

%\noindent \textbf{Challenges}
\hupar{Challenges.}Supervised learning algorithms are powerful in optical flow estimation \cite{dosovitskiy2015flownet, sun2018pwc}. However, frame-based and event-based \cite{zhu2018ev, lee2020spike} networks are not suitable for spike streams due to the difference of their data modalities. Besides, as we discussed in the introduction, the simple solution, i.e., the images reconstructed from a spike stream are used to estimate optical flow,  is unreasonable. For estimating $\mathbf{W}_{t_i,t_j}$ directly from a spike stream, another simple solution is to input the spike stream  $\{S_n\}$  between times  $t_i$ and  $t_j$ as a multi-channel tensor. However, in this way, the length of $\{S_n\}$ increases with the increasing time interval  $t_j - t_i$, which is inconvenient as the input of network. To avoid the dynamic length of the spike stream,  two spike streams in static temporal windows across time $t_i$ and time $t_j$ respectively can be as input. However, the easy input would be affected by motion blur caused by static temporal windows \cite{spikecamera}.
\\\indent
Dataset is vital for the evaluation of methods. However, there is no dataset at present for spike-based optical flow estimation. In fact, it is difficult to build real optical flow datasets for the spiking camera since sensors cannot accurately calculate the optical flow in high-speed scenes. An alternative solution is to synthesize optical flow datasets for the spiking camera. In order to synthesize valuable spike-based optical flow datasets, high-speed scenes in datasets need to be carefully designed to ensure that they can cover all kinds of motion.
%However, at present, there is no proper dataset providing both spike stream and their corresponding optical flow ground truth. Besides, image-based and event-based \cite{zhu2018ev, lee2020spike} networks are not specially designed for spike stream due to the difference of their data representation. A simple way is to input the spike stream  $\{S_n\}$   between times  $t_i$ and  $t_j$   as a spike sequence for estimating  $\mathbf{W}_{t_i,t_j}$. Unfortunately, in this way, the length of  $\{S_n\}$   increases with the increasing time interval  $t_j - t_i$, which is inconvenient as the input of network. To avoid dynamic length of the input, inspired by (3), we can use spike number frame at time $t$ as the representation of spike stream at time $t$ where each pixel records spike numbers in a fixed temporal window across time $t$. Further, we only need to input two spike number frames at time $t_i$  and $t_j$ to a network. However, the spike numbers in a temporal window corresponds to the average brightness during this period (as (3)). When there is a lot of motion in scenes, the average brightness in the temporal window described by (3) can not accurately describe the instantaneous feature of spike stream, i.e., motion blur would be introduced to the representation.
%\section{Proposed Datasets for the spiking camera}
\section{Spiking Optical Flow Datasets}

We synthesize first two large optical flow datasets for the spiking camera (SPIFT and PHM) based on our proposed the spiking camera simulator (SPCS). In this study, we regard SPIFT as the training set and PHM as the test set. Importantly, PHM includes 10 well-designed scenes with various motion which is beneficial to evaluate the generalization of models. The details are shown in Table \ref{tab:addlabel}.
\begin{table}[htbp]
  \centering
  \caption{Detailed information of datasets.}
    \begin{tabular}{lcccc}
    \toprule
            % & Type    & Sample rate & Resolution & Category & Frame   & Optical flow density & Accord with the real world \\
            \tabincell{c}{Dataset} & \tabincell{c}{Category} & \tabincell{c}{Frame} & \tabincell{c}{Flow density} \\
    \midrule
            SPIFT-Train & 100     & 50000   & 100\% \\
            SPIFT-Validation & 10     & 5000   & 100\% \\
            PHM-Test & 10      & 25100       & 100\% \\
    \bottomrule
    \end{tabular}%
  \label{tab:addlabel}%
\end{table}%
%\subsection{The Simulator for the spiking camera}

\subsection{The Spiking Camera Simulator}
\begin{figure}[thbp]
\includegraphics[width=\linewidth]{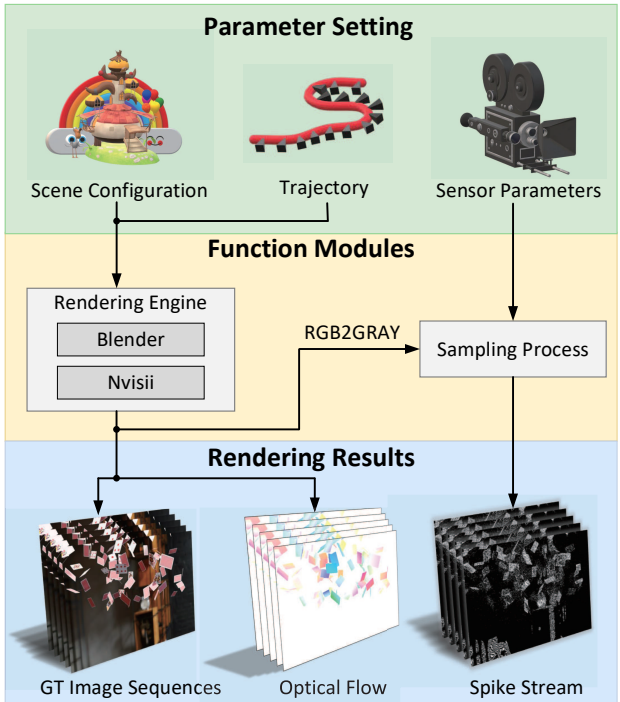}
\centering
\caption{Framework of the spiking camera simulator, SPCS. SPCS relies on a tight coupling with the rendering engines to generate spike streams accurately. Related details are in the supplement.}\label{456}
\end{figure}
%In order to make SCFlow fully learn the ability of extracting optical flow from spike data, The large optical flow dataset for the spiking camera is essential for the generalization of the model.
To generate our datasets, we propose a simulator for the spiking camera (SPCS) coupling with rendering engines tightly. As shown in Fig.~\ref{456}, its architecture mainly consists of parameter setting, rendering engine and sampling process. 
Parameter setting includes adding objects to the scene (scene configuration), generating the trajectory of objects (trajectory), and setting parameters of the spiking camera (sensor parameters).
%2021 10 23
%图4重修 a参数设置标题 b改成运动轨迹小标题 c 换运动轨迹和场景配置的顺序
In SPCS, we provide a one-click generation of parameters, which means that a large number of random scenes can be easily generated.
{Rendering engine} can be used to generate the image sequences sampled by a virtual camera in scenes. SPCS use NVISII\cite{Morrical20nvisii} and Blender \cite{blender} as its engine. Sampling process models the analogue circuit of the spiking camera. When SPCS generates a spike stream, image sequences would be firstly synthesized by rendering engine according to parameter settings, and then the function sampling process would convert image sequences into spike stream. Besides, SPCS also provides noise simulation in the spiking camera. Details are in our supplementary material.
%此外，我们的模拟器也提供了dvs的采样过程，具体细节在附录中。
%模拟器的全部功能在python上实现which允许模拟器很好安装与使用and fig is an example of the output of our simulator。

\begin{figure*}[ht!]
\includegraphics[width=\linewidth]{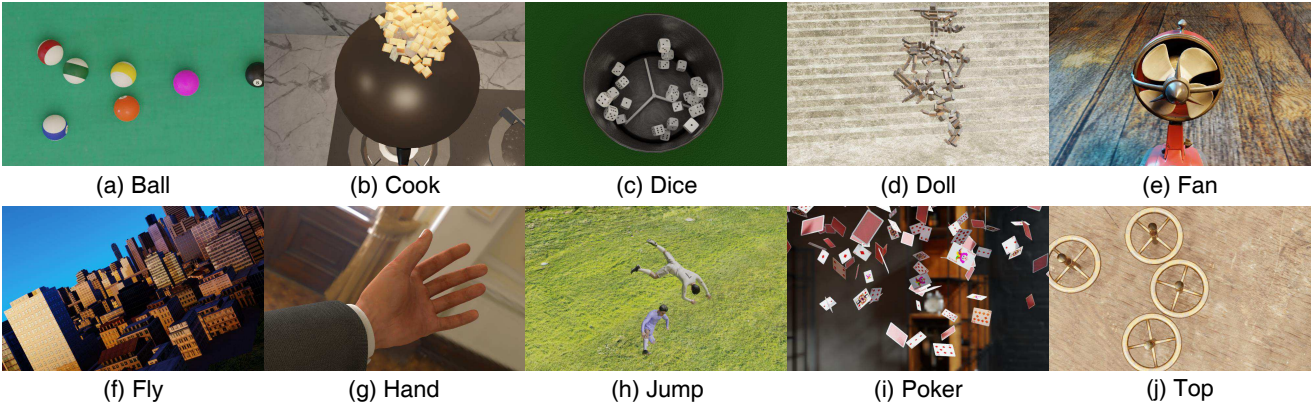}
\centering
\caption{Scenes of Photo-realistic High-speed Motion dataset, denoted as PHM.
}\label{fig:121323}
\end{figure*}

\subsection{SPIFT Dataset}
As shown in Fig.~\ref{fig0}, each scene in the training set describes that different kinds of objects translate and rotate in random background,
% (fig.\ref{fig1}), 
like FlyingChairs \cite{dosovitskiy2015flownet}, and includes spike streams with 500 frames, corresponding GT images and optical flow. Note that we do not generate optical flow at each spike frame, i.e., $\{W_{k \cdot \Delta t, (k + 1) \cdot \Delta t}\}$, $ \Delta t = 1$, $k \in \mathbb{Z}$, since the movement of objects per spike frame is extremely small at the frame rate of spiking cameras (40000Hz). Instead of this, we generate optical flow every 10 spike frames ($\Delta t=10$) and 20 spike frames ($\Delta t=20$) %\lei{Formulas should be avoided in the description paragraphs of the paper. Please replace them with appropriate symbols,($dt = 1, dt = 10, dt=20$), these are unusual. } 
separately from start to end of sampling. We hope the two sets of optical flow can be used to validate the generalization of models on motion with different magnitudes.
% All parameters for scenes (number, size and initial positions of objects) are randomly set to improve diversity (see supplement).
Parameters for scenes are randomly set to improve diversity, more details are included in the supplementary material.

%意义中  on different motion.感觉读起来不顺 还有generalization合理吗？
%11 5 
%注意我们没有每一帧都生成光流因为物体的运动xxx。代替生成{公式}, 我们生成了两组光流数据where第一组光流被生成每间隔10个脉冲帧, i.e.,
%第二组光流被生成。(参考贡献那句话)我们希望这两组光流数据能够被用来validate the generalization of models on motion with different manitude.
%Note that we do not generate optical flow at each spike frame since the movement  of objects per spike frame is extremely small at the frame rate of spiking cameras (40000Hz). Instead of generating $F_{t, t + 1}$, we generate optical flow every 10 spike frames (dt=10) and 20 spike frames (dt=20) separately i.e., $F_{t, t + 10}$ and $F_{t, t + 20}$. we hope the two set of optical flow (dt=10 and dt=20) can validate the generalization of models on motion with different magnitude.

\subsection{PHM Dataset}
Each scene in the testing set is carefully designed and has a lot in common with the real world.

%这11个场景分别对应了不同的
 As shown in Fig.~\ref{fig:121323}, ``Ball" describes that billiard balls collide with each other. ``Cook" describes that the vegetables are stirred in a pot. ``Dice" describes the rotation of dice. ``Doll" describes that some dolls fall from high onto the steps. "Fan" describes fan blade rotation of electric fan. ``Fly" describes that the Unmanned Aerial Vehicle aerial scene. ``Hand" describes that an arm waves in front of the moving camera. ``Jump" describes that two people tumble and jump. ``Poker" describes that pokers are thrown into the air. ``Top" describes that the four tops spin fast and collide.

\section{Method} 
\subsection{Spike Stream Representation}

%光流辅助表征：
%\vsp
%\noindent \textbf{spike stream Representation}
%\vsp
%这里到时候重新仔细过一下 语法这里没有校准 因为是后面改了
\begin{figure}[htbp]
\includegraphics[width=\linewidth]{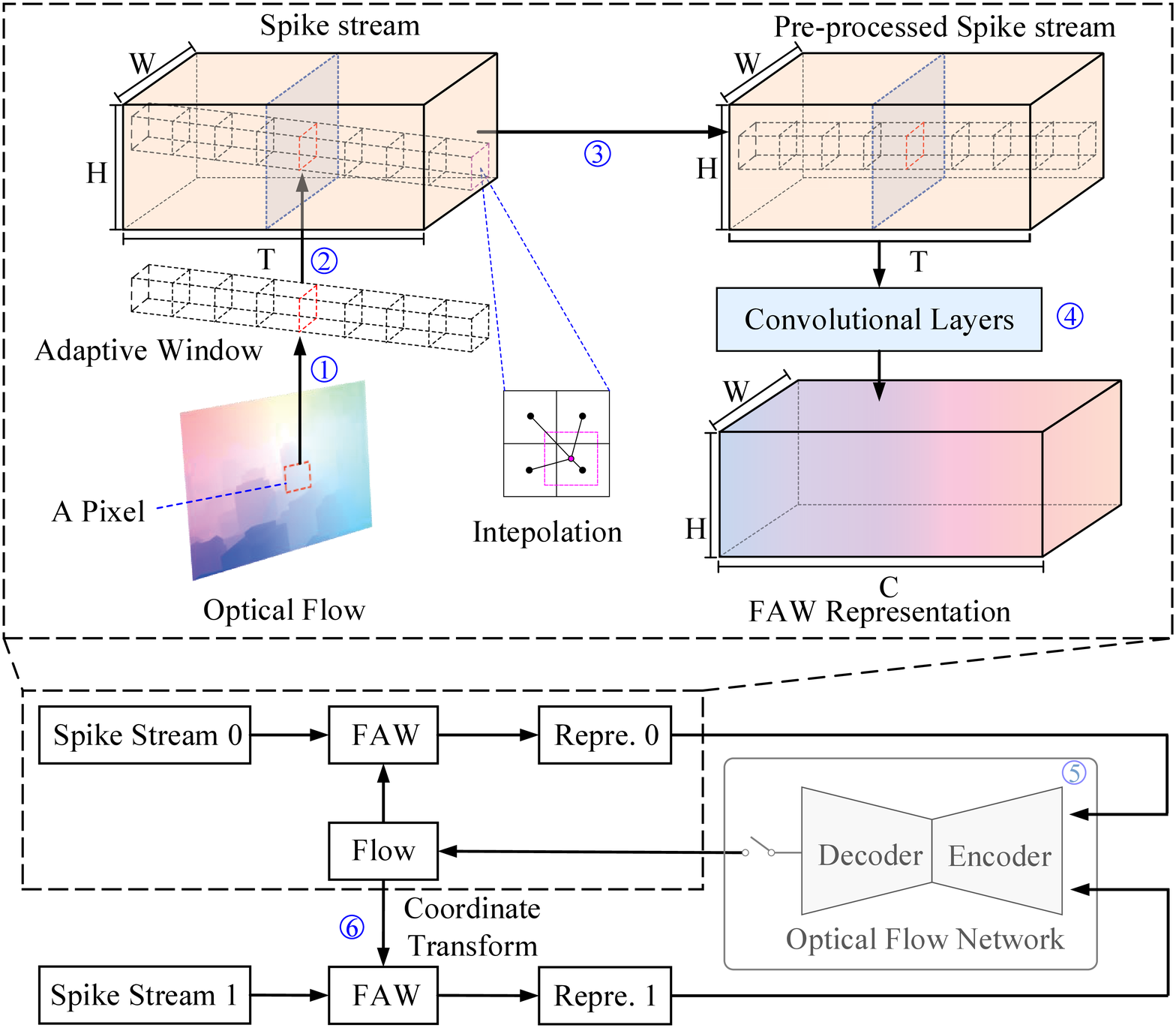}
\centering
\caption{Architecture of proposed representation of a spike stream, FAW. The spike stream includes  $T$   spike frames and each spike frame corresponds to a $H\times W$ binary array. The dimension of obtained FAW is $H\times W \times C$. Spike stream 0 (or 1) corresponds the spike stream across time $t_0$ (or $t_1$).}
\label{fig:FAW}
\end{figure}

\begin{figure*}[htbp]
\includegraphics[width=\linewidth]{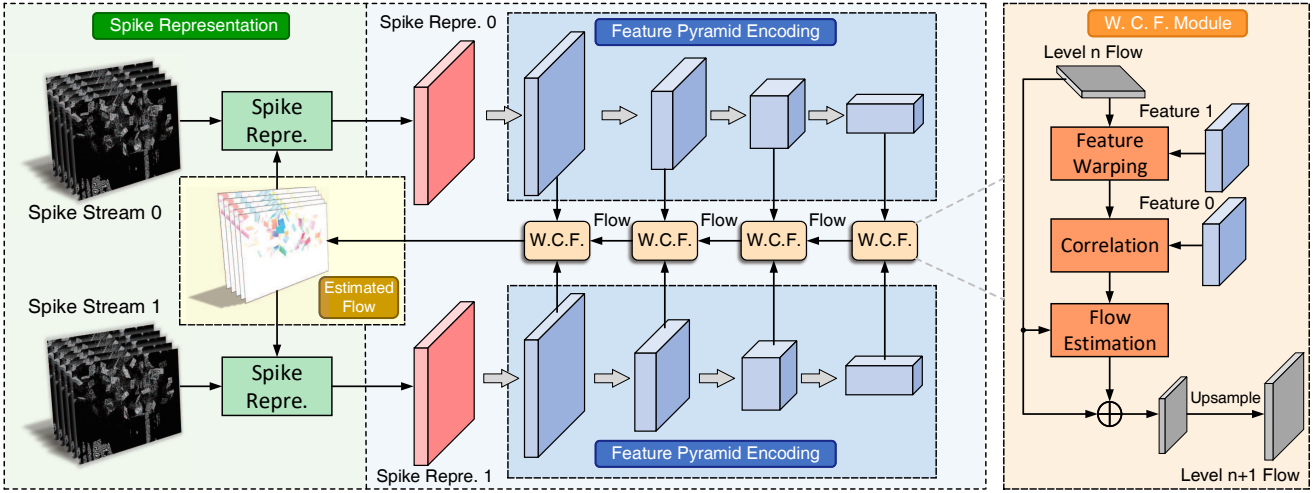}
\centering
\caption{The network architecture of SCFlow. The spike stream is firstly represented by FAW and then constructed to be a feature pyramid. The flow is estimated through the pyramid in a coarse-to-fine manner.}
\label{fig:net}
\end{figure*}

An appropriate input representation of a spike stream for neural networks is crucial. In previous works \cite{rec3,rec4}, the spike stream in a temporal window with fixed direction across time $t_0$ is used as features at time $t_0$. However, the input representation would introduce motion blur due to the static temporal windows. \cite{spikecamera}.
%Although, for each pixel, we can use the spike numbers in a fixed temporal window as the feature of spike stream at time  $t$ and input the spike number frame to network, the spike numbers express the average brightness on the same pixel over a period of time (as (3)). When there are relative motions between the objects and the spiking camera, the brightness on the same pixel is not constant. Hence, the average brightness can not accurately describe the information of spike stream at time  $t$  , i.e., motion blur would be introduced. 

In order to reduce the influence of motion blur on representation at time $t_0$, selected temporal windows should be dynamic and directed. Specifically, if the direction of a temporal window is consistent with the motion trajectory of the pixel, the average brightness in the temporal window would be closer to the brightness of the pixel at time $t_0$.
%现在讲了他们不好，我们要这个做法要解决这个不好，不过没有解释为啥我们这么做能避免这个问题。
To this end, we introduce a proper representation of a spike stream, Flow-guided Adaptive Window (FAW), where prior motion is used as guidance to adaptively adjust the direction of temporal windows for each pixel.
% In FAW, we filter the spike stream along the trajectory of the motion for each pixel. 
For computing FAW at time $t_0$, we firstly use the optical flow from time $t_0$ to time $t_1$ ($t_1 = t_0 + \Delta t$), $W_{t_0, t_1}$ as the prior motion information of all pixels at time $t_0$. Furthermore, we assume that the motion of pixels is a uniform linear motion in a very short time. Hence, the adaptive temporal window of each pixel at time $t_0$ is straight as shown in Fig.~\ref{fig:FAW}\ding{173}, and for the pixel $\mathbf{x} = (x, y)$, the spike information at time $t$ in its adaptive temporal window can be defined as,
\begin{equation}
    \begin{aligned}
    \mathbf{S}_{t_0}^{\rm pre}(\mathbf{x}, t) = \mathbf{S}\Big(\mathbf{x} + \frac{(t - t_0) \cdot {{\mathbf{W}}}_{t_0, t_1}(\mathbf{x})}{\Delta t}, \; t \Big),
    \end{aligned}
\end{equation}
where $t\in[t_0 - \frac{T_{\max} - 1}{2}, t_0 + \frac{T_{\max} - 1}{2}]$, $t_1 = t_0 + \Delta t$ and $\mathbf{S}_{t_0}^{\rm pre}$ is the pre-processed spike stream recording the spike information in all adaptive temporal windows, and $T_{\max}$ is the window length of the input spike stream. Note that spatial locations of the grids in adaptive temporal windows at time $t$ may not be integer, we use the bilinear interpolated spike information in the grids as shown in Fig.~\ref{fig:FAW}\ding{174}. Further, the pre-processed spike stream at time $t_0$ would be encoding into a feature map with multi-channel $\mathbf{S}_{t_0}^{\rm FAW}$ as the FAW at time $t_0$ by two convolutional layers with 32 channels as shown in Fig.\ref{fig:FAW}\ding{175}.
%\revision{The basic usage,representation..blabla of  SCFlow. The detail  of SCFlow will be introduced in Sec.~5.2 ?}.
\\\indent
In fact, there is a key problem, \textit{how to obtain the prior motion (optical flow ${\mathbf{W}}_{t_0, t_1}$)?} As shown in Fig.~\ref{fig:FAW}\ding{176}, we can use the output of Optical Flow Network (see detail of SCFlow in Section 5.2) $\hat{\mathbf{W}}_{t_0, t_1}$ as prior motion. Specifically, during training, we first set prior motion as a zero matrix and a predicted optical flow can be obtained by the forward propagation of SCFlow. Then, we use the predicted optical flow as our prior motion to train SCFlow. During testing, the optical flows that need to be predicted is sorted in chronological order i.e., $\{\mathbf{W}_{k \cdot \Delta t, (k + 1) \cdot \Delta t}\}$, $k \in \mathbb{N}$, $ \Delta t \in \{10, 20\}$. Due to high similarity of motion in adjacent time, the last predicted optical flow is used as a prior motion for the current testing i.e., the prior motion for estimating $\mathbf{W}_{i, i + \Delta t}$ is $\hat{\mathbf{W}}_{i - \Delta t, i}$.
\\\indent
When estimating  ${\mathbf{W}}_{t_0, t_1}(\mathbf{x})$, we only need to input the FAW pair at time $t_0$ and time $t_1$ to our network. 
However, when computing the FAW at time $t_1$, the key frame  $\mathbf{S}(x, t_1)$ and flow $\hat{\mathbf{W}}_{t_0, t_1}$ are on different coordinates. As shown in Fig.~\ref{fig:FAW}\ding{177}, we need to transform $\hat{\mathbf{W}}_{t_0, t_1}$. According to the uniform linear motion assumption for the motion field,  the coordinates transformation to make $\hat{\mathbf{W}}_{t_0, t_1}$ aligned with spike steam at time $t_1$ can be written as,
\begin{align}
&\mathcal{C} \Big( {\mathbf{W}}_{t_0, t_1} (\mathbf{x})\Big) = {\mathbf{W}}_{t_0, t_1} \Big(\mathbf{x} - {\mathbf{W}}_{t_0, t_1}(\mathbf{x})\Big), \\
&\mathbf{S}_{t_1}^{\rm pre}(\mathbf{x}, t) = \mathbf{S}\Bigg(\mathbf{x} + \frac{(t - t_1) \cdot \mathcal{C} \Big( {\mathbf{W}}_{t_0, t_1} (\mathbf{x})\Big)}{\Delta t}, \; t \Bigg),
\end{align}
where $t\in[t_1-\frac{T_{\max}-1}{2}, t_1+\frac{T_{\max}-1}{2}]$, and $\mathcal{C}(\cdot)$ is the coordinate transformation operator.

\begin{table*}[ht]
    \small
    \centering
    \caption{Average end point error comparison with other methods for estimating optical flow on PHM datasets under $\Delta t=10$ and $\Delta t=20$ settings. All methods use spike stream as input and are trained on SPIFT. The best results for each scene and the best average results are marked in bold.}
    \begin{tabular}{clcccccccccccc}
    \toprule
         & Method & Param. & Ball & Cook & Dice & Doll & Fan & Fly & Hand & Jump & Poker & Top & AVG. \\
        \midrule
        \multirow{5}{*}{\rotatebox{90}{ \textbf{$\Delta t=10$} }}
         & EV-FlowNet & 53.43M & 0.567 & 3.030 & 1.066 & 1.026 & 0.939 & 11.072 & 4.558 & 0.824 & 1.306  & 2.625  & 3.501 \\
         & Spike-FlowNet & 13.04M & \textbf{0.500}  & 3.541 & \textbf{0.666} & 0.860 & 0.932 & 11.990 & 4.886 & 0.878 & 0.967 & 2.624 & 3.646 \\
         & RAFT & 5.40M & 0.691 & 2.563 & 1.021 & 0.975 & 0.455 & 10.576 & 3.639 & 0.564 & \textbf{0.842} & 2.614 & 3.162 \\
         & SCFlow-w/oR & 0.57M & 0.597 & 2.185 & 1.288 & 0.606 & 0.464 & 9.625 & 2.551 & 0.370 & 1.269 & 2.602 & 2.883 \\
         & SCFlow (ours) & 0.80M & 0.671 & \textbf{1.651} & 1.190 & \textbf{0.266} & \textbf{0.298} & \textbf{8.783} & \textbf{1.692} & \textbf{0.120} & 1.030 & \textbf{2.166} & \textbf{2.457} \\
        \midrule
        \multirow{5}{*}{\rotatebox{90}{ \textbf{$\Delta t=20$} }}
         & EV-FlowNet & 53.43M & 1.051 & 5.536 & 1.721 & 2.057 & 1.867 & 22.368 & 8.820 & 1.803 & 2.193 & 5.061 & 6.813 \\
         & Spike-FlowNet & 13.04M & \textbf{0.923} & 7.069 & \textbf{1.131} & 1.675 & 1.838 & 25.129 & 9.829 & 1.701 & \textbf{1.373} & 5.257 & 7.385 \\
         & RAFT & 5.40M & 1.267 & 3.905 & 2.182 & 0.546 & 0.689 & 22.550 & 5.021 & 0.300 & 1.414 & 4.330 & 5.926 \\
         & SCFlow-w/oR & 0.57M & 1.321 & 4.493 & 2.601 & 2.206 & 1.083 & 21.419 & 5.654 & 1.159 & 2.320 & 5.143 & 6.346 \\
         & SCFlow (ours) & 0.80M & 1.157 & \textbf{3.430} & 2.205 & \textbf{0.507} & \textbf{0.578} & \textbf{21.127} & \textbf{4.018} & \textbf{0.267} & 1.922 & \textbf{4.327} & \textbf{5.568} \\
         \bottomrule
    \end{tabular}
    \label{tab:final_results}
\end{table*}

\subsection{Network Architecture}
%\revision{With the representaion blabla... , here is our details.... }
As shown in Fig.~\ref{fig:net}, the SCFlow network is designed in a pyramidal encoder-decoder manner. The inputs of the network are the FAW pairs. 
% For predicting optical flow $\mathbf{W}_{t_1, t_2}$, the FAWs $S_{t_1}^{\rm FAW}$ and $S_{t_2}^{\rm FAW}$ are firstly extracted to be feature with 32 channels. 
For each FAW, we build a 4-level feature pyramid  $\{F_i^l(\mathbf{x})\}_{l=1}^{4}, i=0,1$, which denotes the feature for describing the scene at time $t_i$ at $l$-th level. The feature pyramid has 32, 64, 96 and 128 channels in each level respectively.

We estimate the optical flow from higher level to lower level in pyramid. We refer to the well-known PWC-Net \cite{sun2018pwc} to design the decoder. At $l$-th level, we firstly warp  ${F}_1^l(\mathbf{S})$ via the current estimated flow $\hat{\mathbf{W}}_{t_0, t_1}^{l+1}(\mathbf{S})$:
%ask rui: 
\begin{equation}
    {F}_{1,{\rm \hat{\mathbf{W}}_{t_0, t_1}}}^l(\mathbf{x}) = {F}_{1}^l\big(\mathbf{x+\hat{{W}}}_{t_0, t_1}^{l+1}(\mathbf{x})\big),
\end{equation}
where we use bilinear interpolation for the warping operation. We use the features to build a correlation volume \cite{scharstein2002taxonomy, xu2017accurate} to describe the similarity between the features. The correlation volume represents the potential displacements between the two frames, 
% and we normalize the feature in each channel, 
which can be formulated as:
% \begin{equation}
%     \begin{aligned}
%     \footnotesize
%     \mathbf{C}^{l}(\mathbf{x, m}) = \left \langle 
%     \frac{F_{0}^{l}(\mathbf{x}) - \mu_0^l}{\sigma_0^l}, 
%     \frac{F_{1, {\rm w}}^{l}(\mathbf{x + m}) - \mu_{1, {\rm w}}^l}{\sigma_{1, {\rm w}}^l}
%     \right \rangle,
%     \end{aligned}
% \end{equation}
\begin{equation}
    \begin{aligned}
    \footnotesize
    \mathbf{C}^{l}(\mathbf{x, m}) = \left \langle F_{0}^{l}(\mathbf{x}), \;
    F_{1, {\rm w}}^{l}(\mathbf{x + m})
    \right \rangle,
    \end{aligned}
\end{equation}
where $\mathbf{C}^{l}$ represents the correlation volume of the $l$-th feature pyramid, and $\mathbf{m}$   represents the displacement between the two features, $\hat{\mathbf{W}}_{t_0, t_1}$ is written as $\mathbf{w}$ for simplicity and  $\langle \cdot \rangle$   is the channel-wise inner product operation. 
% $\mu$   and  $\sigma$   is the mean and standard deviation value corresponding to feature with the same subscript.

The correlation and the feature extracted from the former spike stream at current level are input to the weight-shared flow estimator. A $1 \times 1$   convolution is employed to adjust the channel numbers at different levels to be 32. The flow estimator consisting of cascaded convolutional layers predicts the residual flow. The refined flow is then upsampled via bilinear kernel as the final output of current level.
The flow is supervised by its ground truth at each level:
\begin{equation}
    \mathcal{L} = \sum_{l=1}^{4} \frac{HW}{2^{4-l}} \Vert \hat{\mathbf{W}}^{l}_{t_0, t_1} - \mathbf{W}^{l}_{t_0, t_1} \Vert_1,
\end{equation}
 %\lei{Please rewrite this formula. .. LOL.}
where $\mathcal{L}$ is our loss function, and $\mathbf{W}^{l}_{t_0, t_1}$ is the ground truth of the flow at $l$-th level.

\section{Experiments}

\begin{figure*}[ht!]
\includegraphics[width=0.96\linewidth]{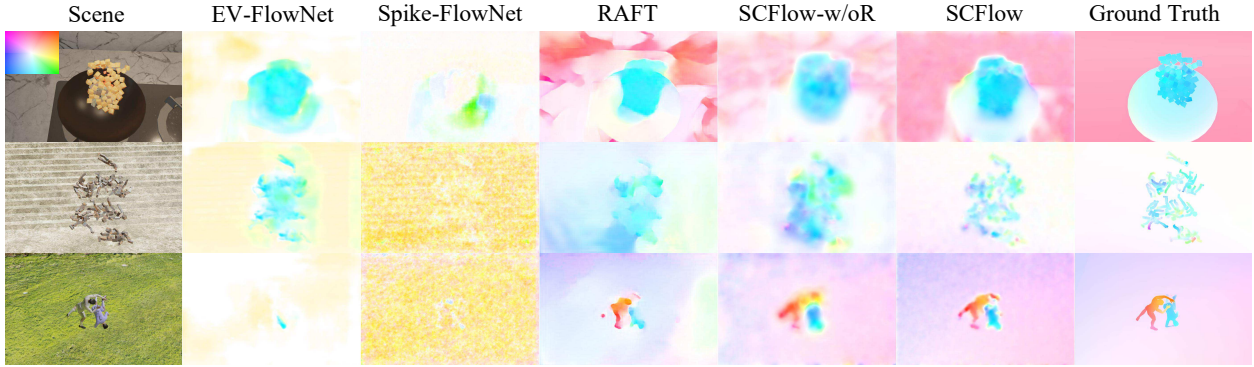}
\centering
\caption{Visual comparison of SCFlow  with other methods under the $\Delta t=10$ setting. On the top left is the color coding of the flow.}
\label{fig:comparison_vis}
\end{figure*}
%\vspace{-0.2cm}
\subsection{Implementation Details}%讲基准方法 以及如何输入 以及训练细节
We train our end-to-end model on the training set of SPIFT in PyTorch. All models are trained by Adam optimizer. We train 40 epochs for the $\Delta t=10$ setting and 80 epochs for the $\Delta t=20$ setting. We randomly crop the input from SPIFT to 480$\times$800 resolution. The batch size in training is set as 4. More details are included in the supplementary material.
\begin{figure*}[htbp]
\includegraphics[width=\linewidth]{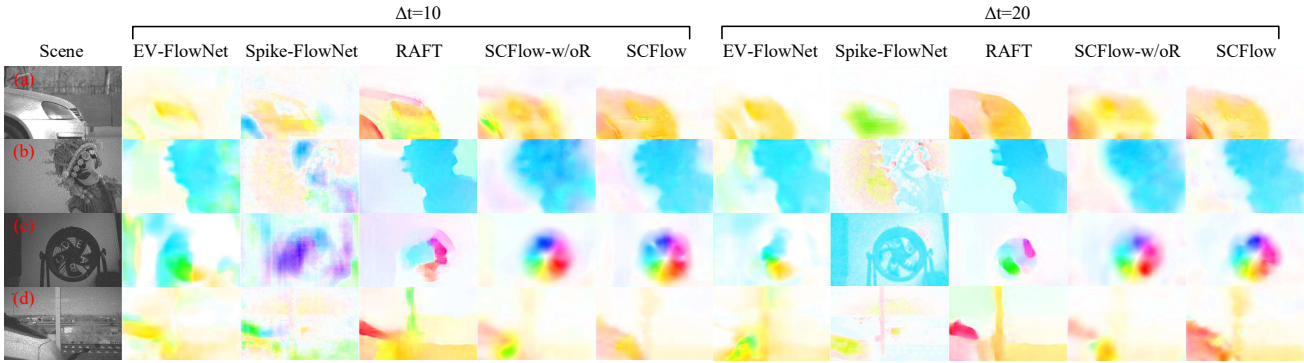}
\centering
\caption{The optical flow visualization on PKU-Spike-High-Speed Dataset \cite{rec1} where $\Delta t= 10$ corresponds optical flow with an interval of 10 spike frames and $\Delta t= 20$ corresponds optical flow with an interval of 20 spike frames. (a) A car traveling at a speed of 100 km/h (kilometers per hour). (b) A free-falling doll.
(c) A high-speed rotating fan.
(d) A train travelling at 350 km/h. All scenes are shot without lens movement.}
\label{fig:real}
\end{figure*}
%\vspace{-0.2cm}
\subsection{Comparison Results}
We compare our method with three categories of methods:
First, we compare our network with neural networks in event-based optical flow, i.e., EV-FlowNet \cite{zhu2018ev} and Spike-FlowNet \cite{lee2020spike}, which are retrained in the same way as SCFlow. The Spike-FlowNet estimates optical flow in a recurrent manner, we split our spike stream to slices with length equals to 2.
Second, we compare our network with state-of-the-art frame-based optical flow network, RAFT \cite{teed2020raft}, which is retrained in the same way with SCFlow. In RAFT, two spike streams in temporal windows across time $t_0$ and time $t_1$ respectively are as input instead of the images at time $t_0$ and $t_1$.
Third, we compare our method with SCFlow without FAW representation (SCFlow-w/oR), which can be viewed as the ablation experiments for FAW.

\hupar{Qualitative Evaluation on Proposed Datasets.}We use average end point error (AEPE) as the evaluation metric, which indicates the average $\ell_2$ norm of the error motion vector between predicted flow and its ground truth. The quantitative comparison results are shown in Table \ref{tab:final_results}. In both $\Delta t=10$ and $\Delta t=20$ settings, our SCFlow gets the best average performance in all methods with a lot fewer parameters. The comparison between SCFlow-w/oR and SCFlow demonstrates that FAW can improve the accuracy of optical flow estimation. The visualization results are shown in Fig.~\ref{fig:comparison_vis}. Evidently, the EV-FlowNet\cite{zhu2018ev}, Spike-FlowNet\cite{lee2020spike}, and RAFT\cite{teed2020raft} cannot estimate optical flow of the spiking camera well, which demonstrates the information processing pattern are not appropriate for spike streams. The flow visualization of SCFlow includes finer motion boundaries and textures than SCFlow-w/oR, which demonstrates the efficiency of the FAW representation.

\hupar{Qualitative Evaluation on Real Data.}In Fig.~\ref{fig:real}, we show  that the visualization results of all methods on real spike streams, PKU-Spike-High-Speed Dataset \cite{rec1}. The visual quality of optical flow produced by our proposed method is evidently better than the competing methods. For RAFT, motion region can be distinguished but predicted direction of motion is inaccurate. 
For Spike-FlowNet and EV-FlowNet, the boundary of predicted motion region is severely blurry and predicted direction of motion is also inaccurate. Besides, by using the FAW representation, predicted motion regions by SCFlow have the more clear boundary than SCFlow-w/oR.
%对于raft，
%图k展示了scflow和其他方法在真实脉冲流(xxx [])visulization results。The visual quality of the reconstructions produced by our proposed method is evidently better than the competing methods. For EV-FlowNet, motion region can be distinguished, but  predicted direction of motion is inaccurate. For Spike-FlowNet, boundary of predicted motion region is severely blurry.  （能说模糊么？ 还是说边界被破坏了) In contrast, our proposed method achieves more reliable optical flow estimation where boundary of predicted motion region is more clear and predicted direction of motion is closer to truth.
\section{Conclusion}
This paper proposed, SCFlow, the first deep learning pipeline for optical flow estimation for the spiking camera. To avoid the motion blur in a spike stream \cite{spikecamera} caused by static temporal windows, a proper input representation of the spike stream, Flow-guided Adaptive Window (FAW). Besides, we synthesize the two optical flow datasets for the spiking camera, SPIFT and PHM, corresponding to random high-speed and well-designed scenes respectively. Finally, we show that SCFlow can predict optical flow from real and synthetic spike streams in different high-speed scenarios.
%\vspace

\hupar{Limitations:} High-speed scenes in low light conditions may be challenging for our model due to more obvious noise and motion blur in spike streams. In future, we plan to fully explore the principles of noise in the spiking camera, extend our datasets and evaluate the performance of our model on scenes in extreme lighting conditions.
%尽管被提出的模型能够端到端的准确地估计光流at 30 fps (see supplement), 它依然是远远达不到实时计算的标准。为了，我们计划优化我们模型的计算效率并且迁移到fpga上。以至于它不能应用到对
%低光照条件下高速运动场景可能是对我们模型的一个挑战因为 脉冲流中有更明显的噪声与运动模糊在这种情况下。In the future, we plan to extend our datasets and evaluate the performance of our model on scenes with extreme lighting conditions.

\section*{Acknowledgments}
This work is supported in part by National Key R\&D Program of China (No. 2020AAA0130400, 2021ZD0109803) and National Natural Science Foundation of China (No. 62072009, 62088102, 62136001, 22127807).
\clearpage
%%%%%%%%% REFERENCES
{\small
\bibliographystyle{ieee_fullname}
\normalem
\bibliography{scflow}
}
\end{document}